\documentclass[11pt]{article}
\usepackage{newunicodechar}
\newunicodechar{↔}{\leftrightarrow}
\usepackage[]{acl}

\usepackage{times}
\usepackage{latexsym}
\usepackage{booktabs}
\usepackage{amsfonts}
\usepackage{amssymb}
\usepackage{tikz}
\usetikzlibrary{arrows.meta, positioning}
\usepackage[T1]{fontenc}

\usepackage[utf8]{inputenc}

\usepackage{microtype}

\usepackage{inconsolata}

\usepackage{graphicx}
\usepackage{amsmath}
\usepackage{multirow}
\usepackage{xcolor}
\usepackage{makecell}

\usepackage{array}

\title{PolyFrame at MWE-2026 AdMIRe 2: When Words Are Not Enough: Multimodal Idiom Disambiguation}

\author{
  \textbf{Nina Hosseini-Kivanani\textsuperscript{1,2}}
\\
\\
  \textsuperscript{1}University of Luxembourg, Luxembourg,
  \textsuperscript{2}Radio Télévision Luxembourg (RTL), Luxembourg
\\
  \texttt{nina.hosseinikivanani@ext.uni.lu}
}

\begin{document}
\maketitle
\begin{abstract}

Multimodal models struggle with idiomatic expressions due to their non-compositional meanings, a challenge amplified in multilingual settings. We introduced PolyFrame, our system for the MWE-2026 AdMIRe~2 shared task on multimodal idiom disambiguation, featuring a unified pipeline for both image+text ranking (Subtask~A) and text-only caption ranking (Subtask~B). All model variants retain frozen CLIP-style vision--language encoders and the multilingual BGE~M3 encoder, training only lightweight modules: a logistic regression and LLM-based sentence-type predictor, idiom synonym substitution, distractor-aware scoring, and Borda rank fusion. Starting from a CLIP baseline (26.7\% Top-1 on English dev, 6.7\% on English test), adding idiom-aware paraphrasing and explicit sentence-type classification increased performance to 60.0\% Top-1 on English, and 60.0\% Top-1 (0.822 NDCG@5) in zero-shot transfer to Portuguese. On the multilingual blind test, our systems achieved average Top-1/NDCG scores of 0.35/0.73 for Subtask~A and 0.32/0.71 for Subtask~B across 15 languages. Ablation results highlight idiom-aware rewriting as the main contributor to performance, while sentence-type prediction and multimodal fusion enhance robustness. These findings suggest that effective idiom disambiguation is feasible without fine-tuning large multimodal encoders.

\end{abstract}

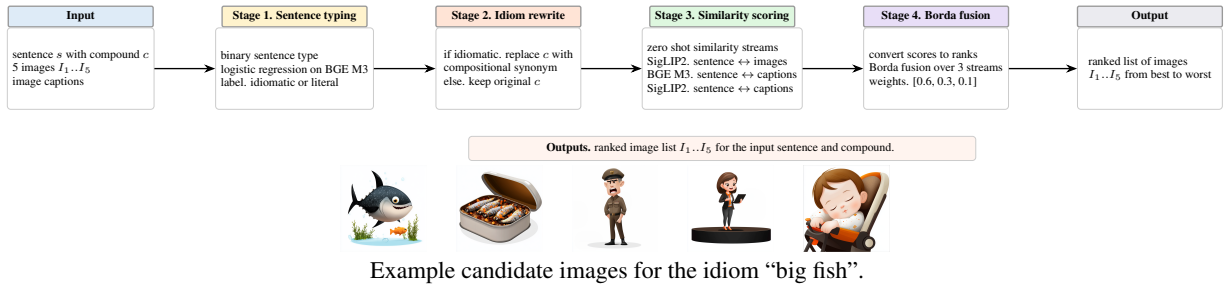
\begin{figure*}[t]
\centering
\resizebox{\linewidth}{!}{%
\begin{tikzpicture}[
  font=\scriptsize,
  >=Stealth,
  node distance=1.4cm,
  header/.style={
    rounded corners=1pt,
    draw=black!40,
    thick,
    inner sep=3pt,
    minimum width=3.0cm,
    align=center
  },
  body/.style={
    rounded corners=2pt,
    draw=black!20,
    inner sep=3pt,
    minimum width=3.0cm,
    minimum height=1.7cm,
    align=left
  }
]

\definecolor{stageblue}{RGB}{222,235,247}
\definecolor{stageyellow}{RGB}{255,243,205}
\definecolor{stageorange}{RGB}{255,228,214}
\definecolor{stagegreen}{RGB}{222,247,223}
\definecolor{stagepurple}{RGB}{232,222,247}
\definecolor{stagegray}{RGB}{235,235,240}

\node[header,fill=stageblue]                      (hIn)  {\textbf{Input}};
\node[header,fill=stageyellow,right=of hIn]       (hS1)  {\textbf{Stage 1. Sentence typing}};
\node[header,fill=stageorange,right=of hS1]       (hS2)  {\textbf{Stage 2. Idiom rewrite}};
\node[header,fill=stagegreen,right=of hS2]        (hS3)  {\textbf{Stage 3. Similarity scoring}};
\node[header,fill=stagepurple,right=of hS3]       (hS4)  {\textbf{Stage 4. Borda fusion}};
\node[header,fill=stagegray,right=of hS4]         (hOut) {\textbf{Output}};

\node[body,below=0pt of hIn] (bIn) {
  sentence $s$ with compound $c$\\
  5 images $I_1..I_5$\\
  image captions
};

\node[body,below=0pt of hS1] (bS1) {
  binary sentence type\\
  logistic regression on BGE M3\\
  label. idiomatic or literal
};

\node[body,below=0pt of hS2] (bS2) {
  if idiomatic. replace $c$ with\\
  compositional synonym\\
  else. keep original $c$
};

\node[body,below=0pt of hS3] (bS3) {
  zero shot similarity streams\\
  SigLIP2. sentence $\leftrightarrow$ images\\
  BGE M3. sentence $\leftrightarrow$ captions\\
  SigLIP2. sentence $\leftrightarrow$ captions
};

\node[body,below=0pt of hS4] (bS4) {
  convert scores to ranks\\
  Borda fusion over 3 streams\\
  weights. [0.6, 0.3, 0.1]
};

\node[body,below=0pt of hOut] (bOut) {
  ranked list of images\\
  $I_1..I_5$ from best to worst
};

\draw[->,thick] (bIn.east)  -- (bS1.west);
\draw[->,thick] (bS1.east)  -- (bS2.west);
\draw[->,thick] (bS2.east)  -- (bS3.west);
\draw[->,thick] (bS3.east)  -- (bS4.west);
\draw[->,thick] (bS4.east)  -- (bOut.west);

\node[
  rounded corners=2pt,
  draw=black!20,
  fill=stageorange!40,
  inner sep=4pt,
  align=left,
  below=0.55cm of bS3,
  minimum width=10.5cm
] (outputs) {
  \textbf{Outputs.} ranked image list $I_1..I_5$ for the input sentence and compound.
};

\end{tikzpicture}%
}

\vspace{0.08em}
\begin{minipage}{0.85\linewidth}
\centering
\includegraphics[height=1.13cm]{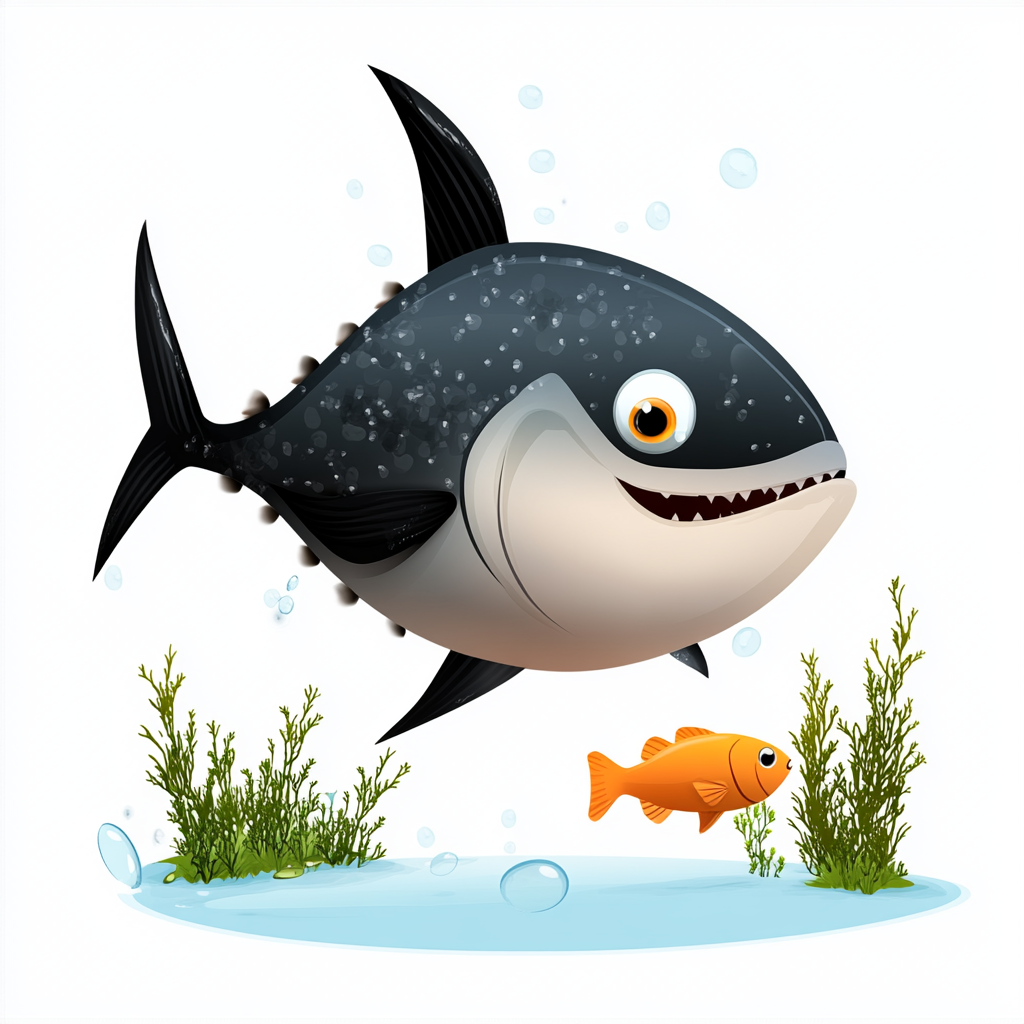}
\hspace{0.2cm}
\includegraphics[height=1.13cm]{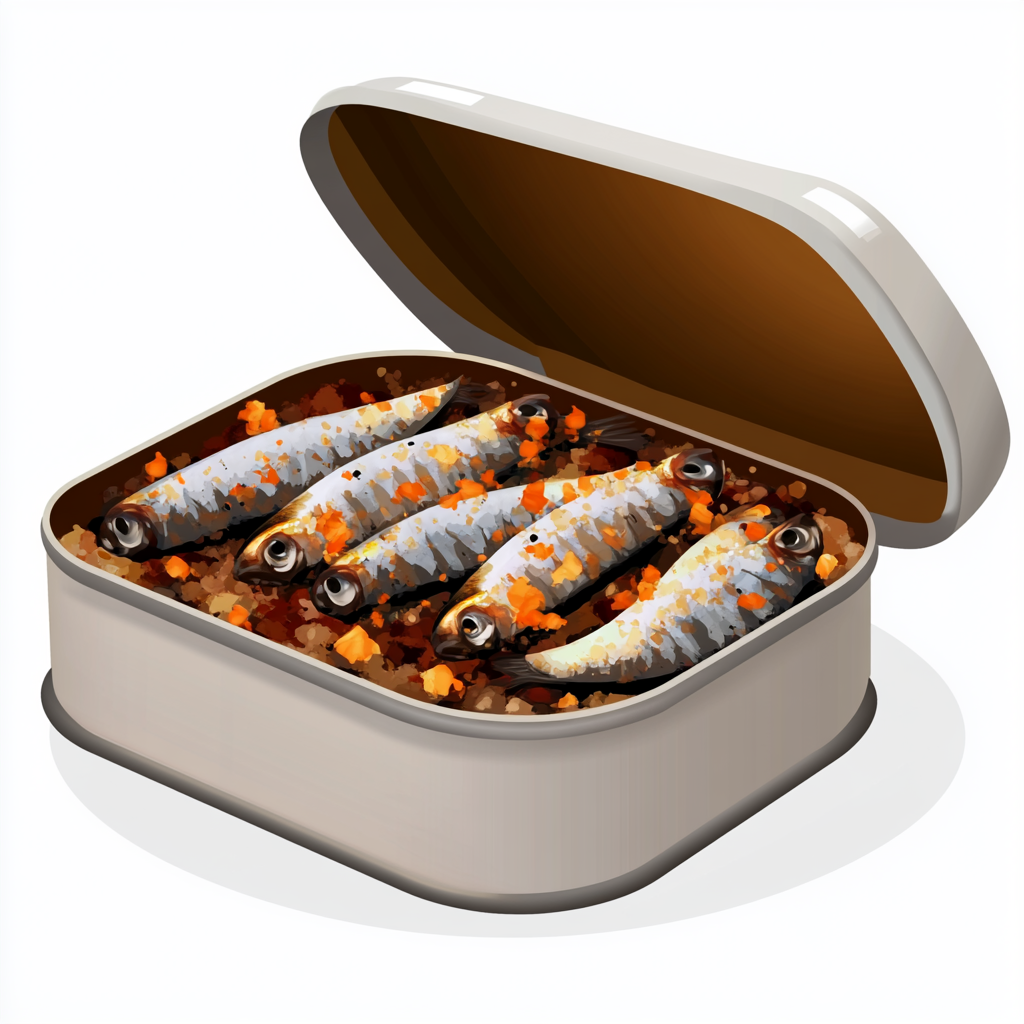}
\hspace{0.2cm}
\includegraphics[height=1.13cm]{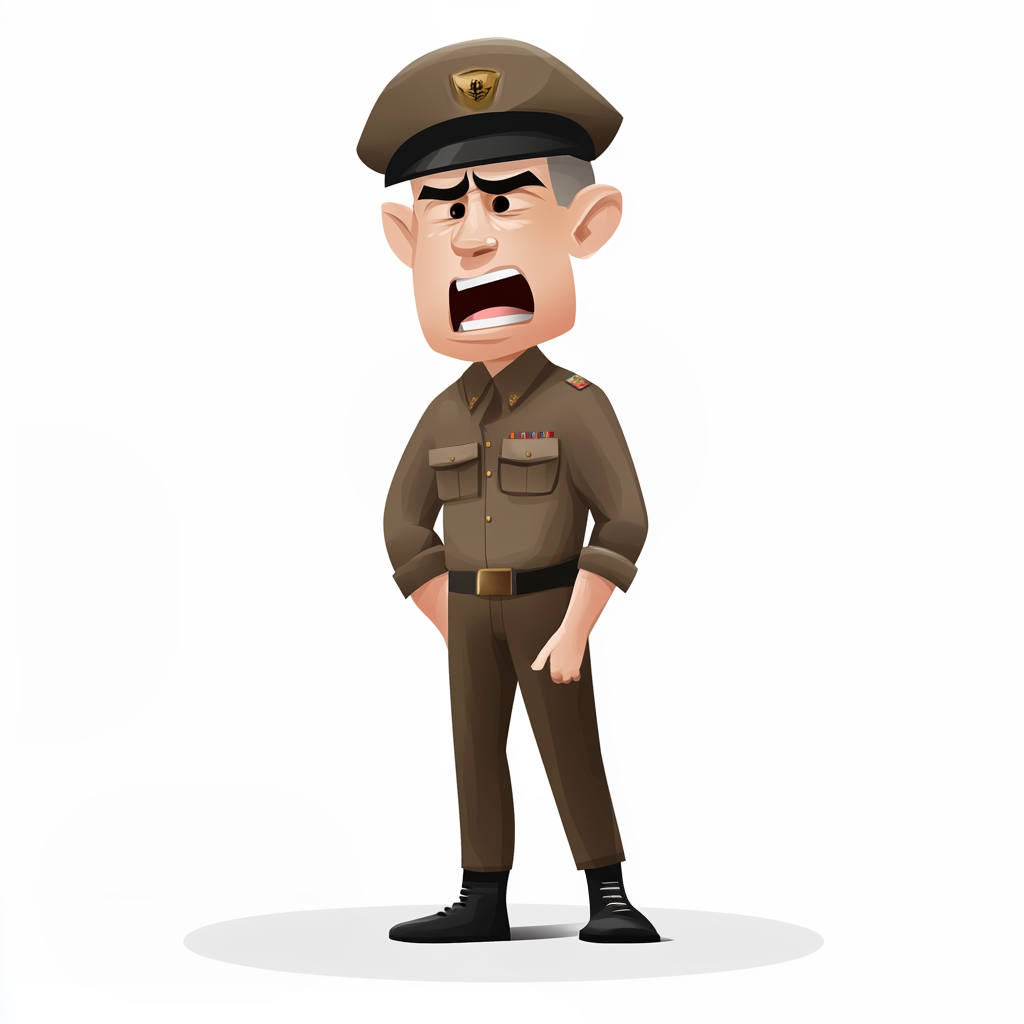}
\hspace{0.2cm}
\includegraphics[height=1.13cm]{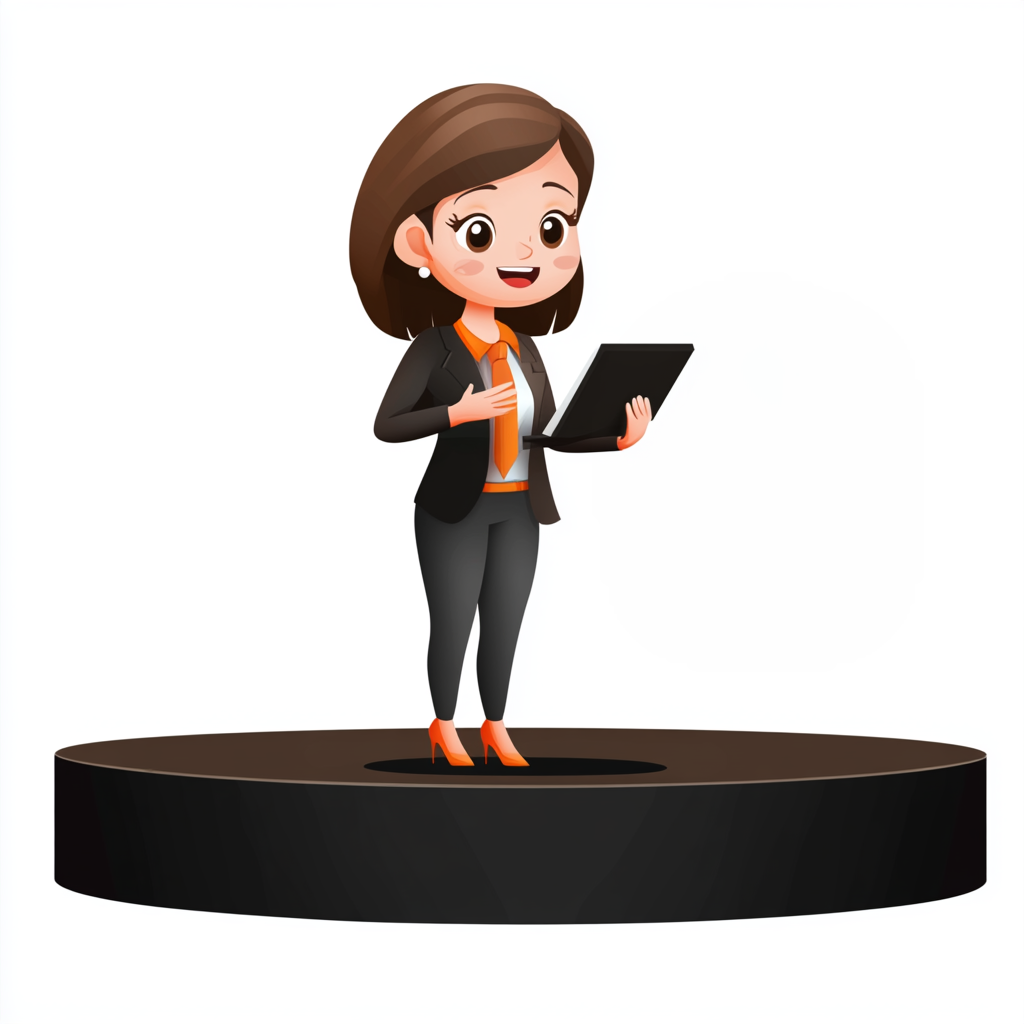}
\hspace{0.2cm}
\includegraphics[height=1.13cm]{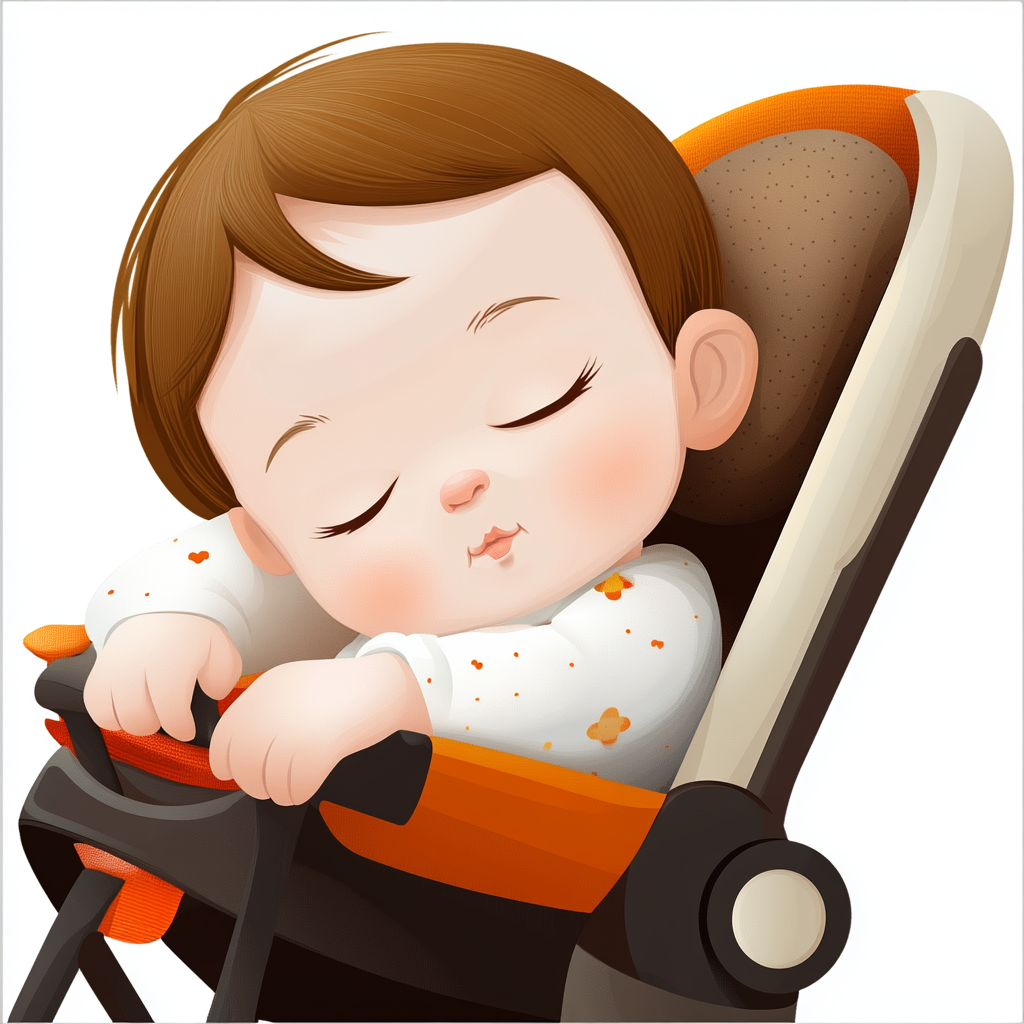}

\footnotesize{Example candidate images for the idiom ``big fish''.}
\end{minipage}

\caption{Overview of the final \textsc{PolyFrame} pipeline. Sentence typing via logistic regression, idiom replacement for idiomatic cases, three zero shot similarity streams with SigLIP2 and BGE M3, and Borda fusion}
\label{fig:admire_pipeline}
\end{figure*}

\section{Introduction}

Idiomatic expressions are a long-standing problem for computational semantics because their figurative meanings are not compositionally predictable from surface form~\cite{flor2025survey,zeng2021idiomatic}. Models trained primarily on literal data often interpret idioms word by word, which leads to semantic drift in downstream tasks such as translation, retrieval, and captioning~\cite{pickard-etal-2025-semeval}. This effect is amplified in multilingual settings, where idioms differ in lexicalisation and transparency across languages~\cite{domhan2022devil,cap2015account}. Recent work treats idiomaticity as a structured prediction problem, for example, by classifying usages as literal or idiomatic or by ranking contextually appropriate interpretations~\cite{endalie2023deep}. 

AdMIRe~2.0 provides sentences containing potentially idiomatic nominal compounds together with five candidate images and parallel caption-based variants~\cite{arslan-2026-admire}. Subtask~A asks systems to rank images according to how well they reflect the intended meaning of the compound in context. Subtask~B replaces images with captions and prompts to probe text-only reasoning under the same schema. Both subtasks implicitly require robust sentence-type judgements (idiomatic vs.\ literal) and fine-grained alignment of the chosen sense with visual or textual cues.

This paper presents our system description for AdMIRe~2.0~\cite{arslan-2026-admire}. We build a shared data and evaluation pipeline and instantiate three variants within a single architecture: (i) a CLIP-based multimodal baseline for Subtask~A; (ii) an improved multimodal ranker with supervised sentence-type classification (logistic regression plus a literal-first LLM classifier), idiom synonym replacement, distractor-aware scoring, and fusion of visual and caption-based scores; and (iii) a text-only counterpart of the improved ranker for Subtask~B (Figure~\ref{fig:admire_pipeline}). All systems use frozen CLIP-style vision--language encoders and the multilingual BGE~M3 sentence encoder~\cite{chen-etal-2024-m3}, with learning restricted to lightweight idiomaticity classifiers, scalar fusion weights, and rank-level Borda ensembles over multiple CLIP and LLM backends. Our experiments examine how sentence-type prediction, idiom-aware rewriting, and multimodal fusion jointly improve idiom-sensitive image and caption ranking across languages.

\section{Related Work}

\textbf{Idiom processing and idiomaticity detection.}
Text-based idiom processing has been studied extensively through shared tasks and benchmarks that ask models to decide whether a target expression is used literally or idiomatically in context (e.g.,~\cite{jakhotiya2022jarvix,boisson2022cardiffnlp}). SemEval~2022 Task~2 introduced multilingual idiomaticity detection for verbal multiword expressions and showed that glosses, translations, and lexical knowledge bases can improve cross-lingual disambiguation~\cite{phelps2022sample}. Subsequent work has evaluated large language models on idiom detection and translation, reporting that they still over-predict idiomatic readings in literal contexts and require careful prompting to approach the performance of supervised baselines~\cite{phelps2024sign}. Other studies have proposed targeted metrics for evaluating idiom translation in neural machine translation and documented a strong tendency toward literal renderings~\cite{khan2025evaluating}. The AdMIRe shared tasks extend this line of work by framing idiomaticity as a multimodal ranking problem and by providing graded relevance annotations for sentence type and image suitability~\cite{torunoğluselamet2026parallelcrosslingualbenchmarkmultimodal,pickard2025semeval}.

\textbf{Multimodal idiom understanding and vision--language models.}
Multimodal encoders such as CLIP learn aligned image--text representations via large-scale contrastive learning and have become standard backbones for zero-shot classification and retrieval. However, their pretraining data and objectives are not tailored to figurative language, and idiomatic usages are often under-represented. Recent approaches adapt CLIP to idiom interpretation by enriching prompts with natural-language definitions, few-shot examples, or idiomatic paraphrases, and by combining visual similarity with textual sentence embeddings~\cite{markchom2025uor,wang2025mchirc}. AdMIRe~2.0 offers a test bed for such strategies by providing both image-based and caption-based variants of the same task. Our systems follow this direction: we keep OpenCLIP-style encoders frozen, pair them with BGE~M3 for multilingual sentence embeddings, and focus learning capacity on lightweight idiomaticity classifiers and fusion mechanisms that combine visual, caption-based, and definition-based scores within a single architecture.

\section{Methodology\protect\footnote{https://github.com/NinaKivanani/PolyFrame}}

We describe our systems for the two AdMIRe~2.0 subtasks, which jointly probe sentence-level idiomaticity and multimodal disambiguation (see Figure~\ref{fig:admire_pipeline} for an overview). Subtask~A presents a context sentence containing a potentially idiomatic nominal compound and five candidate images; the system must rank the images by appropriateness. Subtask~B reuses the same TSV schema but replaces images with captions and prompts, yielding a text-only ranking task. Sentence-type prediction is evaluated by accuracy, and ranking by Top-1 accuracy and NDCG@5.

\subsection{Task setup and system variants}

For Subtask~A, the organisers provide supervised splits for English (EN) and Portuguese (PT) with train, development, test, and cross-evaluation partitions. Each instance consists of a sentence, the target compound, five image file names organised under idiom-specific directories, and optionally image captions and prompts. For evaluation, we run our system on the official blind-test bundles covering 15 languages, resolving image paths via a language-aware global root and tracking the number of processed instances per language (approximately 50--360) to ensure full coverage.

For Subtask~B, we reuse the TSV structure but operate in text-only mode: image identifiers are retained for compatibility, while ranking uses captions and prompts only. 
Within a shared architecture, we instantiate three system variants to study the impact of modality, sentence-type modelling, and idiom replacement: (i) a CLIP-based multimodal baseline for Subtask~A, (ii) an improved multimodal ranker with explicit sentence-type classification, idiom replacement, and distractor-aware fusion, and (iii) a text-only variant of the improved ranker for Subtask~B.

\subsection{Representations and encoders}

Each TSV row is parsed into a context sentence $x$, a target compound $c$, five image identifiers $\{i_1,\dots,i_5\}$, and, when available, image captions $\{c_k\}_{k=1}^{5}$. Supervised splits additionally provide a gold sentence type $y_{\text{sent}} \in \{\text{idiomatic}, \text{literal}\}$ and a gold ranking $\pi^* = [\pi^*_1, \dots, \pi^*_5]$.

All systems use vision-language encoders for multimodal understanding. The baseline uses a multilingual CLIP variant (xlm-roberta-large-ViT-H-14) with frozen LAION-5B weights. The improved systems employ a dual-encoder architecture: SigLIP2 (ViT-SO400M-14-SigLIP2-378)~\cite{tschannen2025siglip}  as the primary vision-language encoder for image-text similarity, combined with BGE-M3 as a separate multilingual text encoder for text-only ranking via caption matching. Rankings from both encoders are fused using Borda-count aggregation. Additional CLIP or SigLIP models can be added in ensemble mode with weighted fusion.

For sentence and caption representations we additionally use BGE~M3, a multilingual dense retrieval encoder. Given a sentence--compound pair $(x,c)$, we concatenate them and obtain an embedding $g(x,c) \in \mathbb{R}^d$, which provides text-only similarity scores for caption-based ranking and serves as the feature space for a lightweight binary sentence-type classifier.

\subsection{Sentence-type prediction and ranking}

\textbf{Sentence-type prediction.}
Sentence-type prediction combines a supervised classifier and an LLM-based component. We train a two-way logistic regression model on EN data using BGE~M3 features $g(x,c)$ to predict $y_{\text{sent}} \in \{\text{literal}, \text{idiomatic}\}$. This serves as the primary sentence-type signal when available.

As a complementary mechanism and cross-lingual fallback, we use literal-first prompting with external LLMs (GPT-4o, Qwen3-32B, Llama~3.1-70B, DeepSeek-v3, Mistral). For each compound–language pair, the system first asks the LLM to generate a small set of clearly literal example sentences, which are cached. The final prompt presents these examples together with the target sentence and instructions to decide between \textsc{literal} and \textsc{idiomatic}. We allow brief reasoning but constrain the output to a single label. 

\textbf{Baseline multimodal ranker.}
The baseline first predicts sentence type heuristically, then constructs prompts, and finally performs visual and textual ranking. Given the predicted type, we build a small set of text queries that combine the sentence, the compound, short natural-language definitions from a hand-curated idiom lexicon, and few-shot examples. These queries are encoded with the CLIP text tower and averaged to obtain a single query embedding $q$.

For visual ranking, each image $i_k$ is mapped to an embedding $v_k = f_\text{img}(i_k)$, and similarity scores are computed via cosine similarity between $q$ and $v_k$, optionally scaled by a temperature $\tau$ to sharpen the distribution over the five candidates; images are then ranked by descending similarity. Caption-based ranking is analogous, using captions instead of images. Visual and caption-based rankings are fused via a Borda-style scheme that converts rank positions into scores and interpolates them, typically with a higher weight on visual information.

\textbf{Improved multimodal ranker.}
The improved ranker extends the baseline with supervised sentence-type classification, idiom synonym replacement, and distractor-aware fusion. For sentences classified as idiomatic, we rewrite the context by replacing occurrences of $c$ with a compositional paraphrase that makes the figurative meaning explicit; the modified sentence is then used to construct CLIP prompts, while literal sentences are left unchanged.

The final ranking fuses three score streams: (1) SigLIP2 vision-language similarity (sentence $\leftrightarrow$ images), (2) BGE-M3 text similarity (sentence $\leftrightarrow$ captions), and (3) SigLIP2 text similarity (sentence $\leftrightarrow$ captions via SigLIP2's text encoder). In our submitted configuration we use weighted Borda-count aggregation with fixed weights of [0.6, 0.3, 0.1] for vision-language similarity, BGE-M3 text similarity, and SigLIP2 caption similarity, respectively, in image+text mode. For text-only mode, weights are adjusted to [0.0, 0.7, 0.3] to focus on text components while ignoring vision. A confidence measure derived from the gap between the top two scores in each stream can adjust these weights. For non-English languages, an optional cross-lingual mode combines scores computed on the original sentence with scores computed on an English translation obtained from the same LLM used for literal-first classification.

\textbf{Text-only ranker.}
The text-only system reuses the same architecture in caption-only mode. Image embeddings are set to zero and receive zero weight in the fusion stage; sentence-type prediction, idiom synonym replacement, and caption-based scoring with BGE~M3 and SigLIP2 remain unchanged. 
This design ensures that improvements over random ranking are attributable purely to text modelling of captions and that ablations of the classifier, literal-first prompting, or idiom replacement are directly comparable across subtasks.

\subsection{Ensembles, transfer strategies}

\textbf{Model ensembles and Borda fusion.}
To improve robustness and support zero-shot transfer, we employ ensembles at both encoder and classifier level with a unified rank-level fusion scheme. On the vision side, we optionally use an ensemble of CLIP variants. Our best-performing configuration uses a single SigLIP2 model (ViT-SO400M-14-SigLIP2-378) rather than an ensemble, as this provided optimal performance while maintaining efficiency. The system supports ensemble configurations where multiple CLIP or SigLIP variants can be combined using weighted Borda fusion, but the submitted system operated in single-model mode.

For sentence-type prediction, we evaluate several LLMs under the same literal-first prompting strategy; in the final configuration, we combine Qwen3-32B and GPT-4o with weights $[0.6, 0.4]$, selected based on development performance, while other LLMs act as fallbacks. In all cases, we apply weighted Borda fusion: each candidate’s rank in each list is converted into a score (higher for better ranks), multiplied by the corresponding fusion weight, and summed. Candidates are then re-ranked by their aggregated Borda scores, providing a consistent mechanism for combining both modalities and model variants.

\textbf{Zero-shot and few-shot transfer.}
To handle idioms and languages with limited training data, we support both zero-shot and few-shot strategies. When few-shot prompting is enabled, CLIP queries are augmented with hand-curated examples that illustrate literal and idiomatic usage (e.g., idiomatic \emph{big fish} in a company vs.\ literal \emph{big fish} in a fishing context), which are prepended to CLIP text queries to guide the vision--language encoder. For languages without training data, we rely on the multilingual capabilities of SigLIP2 and BGE~M3: the system applies zero-shot classification directly to non-English sentences using the multilingual encoders, and when this fails, the system falls back to simple heuristics based on compound frequency in captions and basic lexical markers. This zero-shot setup yields reasonable performance (around 60\% accuracy on PT) without language-specific training. 



\section{Results \& Discussion}

We evaluate our systems using the official AdMIRe~2.0 metrics, reporting Top-1 accuracy and NDCG@5. Experiments cover both subtasks and include baseline comparisons, strategy-specific variants, ablation studies, and the official blind-test evaluation on Codabench. We used a temperature of $\tau = 0.7$ for similarity scaling and Borda-count-based fusion as the default for final evaluation.

The CLIP-based baseline, which relies on frozen vision--language embeddings and heuristic sentence typing, achieves 26.7\% Top-1 accuracy on the EN development set and 6.7\% on the test set (see Table~\ref{tab:dev-test-results}). This large drop highlights the difficulty of the task and confirms that naive multimodal retrieval is strongly biased toward literal interpretations when idiomatic compounds are present.

\begin{table}[hpt!]
\centering

\resizebox{\columnwidth}{!}{%
\begin{tabular}{lcccccc}
\toprule
\multirow{2}{*}{\textbf{System}} 
  & \multicolumn{2}{c}{\textbf{EN dev}} 
  & \multicolumn{2}{c}{\textbf{EN test}} 
  & \multicolumn{2}{c}{\textbf{PT dev (zero-shot)}} \\
\cmidrule(lr){2-3} \cmidrule(lr){4-5} \cmidrule(lr){6-7}
  & Top-1 (\%) & NDCG@5 
  & Top-1 (\%) & NDCG@5 
  & Top-1 (\%) & NDCG@5 \\
\midrule
CLIP baseline          & 26.7 & 0.655 &  6.7 & 0.607 &  --  &  --   \\
+ Idiom replacement    & 60.0 & 0.800 &   -- &   --  &  --  &  --   \\
+ LLM sentence typing  & 40.0 & 0.739 &   -- &   --  &  --  &  --   \\
+ All improvements     & 60.0 & 0.797 &   -- &   --  & 60.0 & 0.822 \\
\bottomrule
\end{tabular}%
}
\caption{System performance across evaluation sets.}
\label{tab:dev-test-results}
\end{table}

Introducing idiom synonym replacement yields the largest single improvement, raising Top-1 accuracy on the EN development set to 60.0\% (Table~\ref{tab:dev-test-results}). This confirms observations from AdMIRe~2024 that rewriting idiomatic expressions into compositional paraphrases substantially reduces literal bias in vision--language models. Literal-first sentence classification using an external LLM improves performance to 40.0\% when used in isolation, indicating that sentence-type awareness is beneficial but insufficient without explicit paraphrasing. The best submission didn't use LLM classification. Combining sentence-type prediction, idiom replacement, and multimodal fusion in the improved ranker maintains 60.0\% Top-1 accuracy while providing greater robustness across prompt conditions and idiom types. Zero-shot transfer to PT also reaches 60.0\% Top-1 and 0.822 NDCG@5, suggesting that the core strategies generalise across related languages.

\begin{table}[t!]
\centering
\scriptsize
\setlength{\tabcolsep}{10pt}
\begin{tabular}{lccc}
\toprule
\textbf{System}        & \textbf{Top$_1$A} & \textbf{Top$_1$B} & \textbf{ST Acc.} \\
\midrule
\makecell[l]{CLIP baseline\\(XLM-RoBERTa-ViT-H-14)} 
              & 26.7 & 6.7 & 53.3 \\
\midrule
GPT-4o        & \textit{32.5} & \textit{28.0} & \textit{58.0} \\
Qwen3-32B     & \textit{31.0} & \textit{26.5} & \textit{56.5} \\
Llama3.1-70B  & \textit{29.5} & \textit{25.0} & \textit{55.0} \\
DeepSeek-v3   & \textit{28.0} & \textit{23.5} & \textit{53.5} \\
Mistral-large & \textit{30.0} & \textit{25.5} & \textit{54.5} \\
\midrule
\makecell[l]{\textbf{Ensemble}\\\textbf{(Qwen3+GPT-4o)}} 
              & \textbf{\textit{33.5}} & \textbf{\textit{29.0}} & \textbf{\textit{59.0}} \\
\midrule
\textbf{Best submission} 
              & \textbf{35.0} & \textbf{32.0} & \textbf{--} \\
\bottomrule
\end{tabular}
\caption{Multilingual blind test results (15 languages). Best submission used SigLIP2 + idiom replacement with zero-shot classification.}
\label{tab:main-results}
\end{table}

\textbf{Multilingual blind-test results.}
We submitted our systems to the official Codabench evaluation for both subtasks. On the multilingual blind test covering 15 languages, our submission achieved an average Top-1 accuracy of 0.35 and an NDCG of 0.73 for Subtask~A, and 0.32 Top-1 accuracy with an NDCG of 0.71 for Subtask~B. These scores correspond to a mid-range ranking among participating systems at the time of evaluation. Performance was consistent across languages, with comparable scores on Chinese and non-Chinese subsets, indicating stable cross-lingual behaviour in a fully zero-shot setting (Table~\ref{tab:main-results}).

For the blind-test phase, we apply the improved system unchanged to all 15 evaluation languages, using the official directory structure and file formats. The submitted configuration combines SigLIP2 with idiom replacement and zero-shot sentence-type classification. We verify full coverage for each language, with sample counts ranging from 48 to 363 instances (see Table~\ref{tab:per_language_subtasks} in the Appendix).

\textbf{Ablation studies.}
Ablation experiments on the EN development set show that idiom replacement accounts for most of the performance gains, while sentence-type prediction mainly improves stability and reduces variance across idioms. Multimodal fusion consistently outperforms unimodal ranking, and Borda-style aggregation is more robust than alternative rank-combination strategies across hyperparameter settings. Temperature and fusion-weight sweeps show limited sensitivity around the chosen defaults, indicating that the improvements are not driven by fragile tuning.


Taken together, the results suggest that most of the attainable gains come from better framing of the vision–language matching problem rather than from stronger back-end language models~\cite{radford2021learning}. Idiom rewriting and multimodal fusion close much of the gap between the CLIP baseline and our best systems~\cite{gao2024clip}, while LLM-based sentence-type prediction yields smaller incremental gains at the cost of additional latency, API dependence, and prompt sensitivity~\cite{jin2025efficient}. This trade-off is reflected in the final Codabench submission, which uses SigLIP2 with idiom replacement and zero-shot sentence typing, but omits LLM classification despite its slightly higher scores on the EN development set. The relatively modest drop from Subtask A to Subtask B and the stable NDCG values across languages indicate that caption-based reasoning can approximate image-based disambiguation when captions are informative, but also highlight persistent weaknesses on low-resource and culturally marked varieties such as Spanish Ecuador and Uzbek~\cite{tschannen2025siglip}. Overall, the pattern of ablation results supports our design choice to prioritise frozen encoders and lightweight, transparent adaptations over heavy fine-tuning or tightly coupled LLM components~\cite{xing2024survey}.

\section*{Limitations and Future Work}
A key limitation of our approach is that the idiom synonym replacement database is manually curated and currently covers only around 50 common English idioms, with limited coverage for the 14 non-English languages in the blind test. This restricts the effectiveness of idiom replacement for less frequent expressions and for many non-English instances. Future work will explore learned fusion weights as an alternative to our current fixed Borda-based scheme. Since our experiments so far rely mainly on decoder-style large language models, we also plan to incorporate encoder-based models (e.g., BERT) and encoder–decoder architectures (e.g., T5, mBART). These architectures may capture idiomaticity in different ways and provide complementary perspectives on representation learning for figurative language.

\section*{Acknowledgments}
This work was carried out within the project ``Comprehensive Text-to-Speech Development for Luxembourgish with Emotional Enhancements (LuxVoice),'' project reference 19205922. The present project was supported by the National Research Fund, Luxembourg.

\bibliography{custom}

@inproceedings{arslan-2026-admire,
    title = {{MWE-2026 Shared Task 2: AdMIRe 2 - Advancing Multimodal Idiomaticity Representation}},
    author = {Arslan, Do{\u{g}}ukan and Wilkens, Rodrigo and He, Wei and Toruno{\u{g}}lu-Selamet, Dilara and Pickard, Thomas and Villavicencio, Aline and Pagano, Adriana S.  and Eryi{\u{g}}it, G{\"u}l{\c{s}}en},
    editor = {Ojha, Atul Kr.  and
      Mititelu, Verginica Barbu  and
      Constant, Mathieu  and
      Do{\u{g}}ru{\"o}z, A. Seza  and
      Rademaker, Alexandre and
      Stoyanova, Ivelina},
    booktitle = "Proceedings of the 22nd Workshop on Multiword Expressions (MWE 2026)",
    month = march,
    year = "2026",
    address = "Rabat, Morocco",
    publisher = "Association for Computational Linguistics"
}

@inproceedings{chen-etal-2024-m3,
    title = "{M}3-Embedding: Multi-Linguality, Multi-Functionality, Multi-Granularity Text Embeddings Through Self-Knowledge Distillation",
    author = "Chen, Jianlyu  and
      Xiao, Shitao  and
      Zhang, Peitian  and
      Luo, Kun  and
      Lian, Defu  and
      Liu, Zheng",
    editor = "Ku, Lun-Wei  and
      Martins, Andre  and
      Srikumar, Vivek",
    booktitle = "Findings of the Association for Computational Linguistics: ACL 2024",
    month = aug,
    year = "2024",
    address = "Bangkok, Thailand",
    publisher = "Association for Computational Linguistics",
    url = "https://aclanthology.org/2024.findings-acl.137/",
    doi = "10.18653/v1/2024.findings-acl.137",
    pages = "2318--2335"
}

@article{tschannen2025siglip,
  title={Siglip 2: Multilingual vision-language encoders with improved semantic understanding, localization, and dense features},
  author={Tschannen, Michael and Gritsenko, Alexey and Wang, Xiao and Naeem, Muhammad Ferjad and Alabdulmohsin, Ibrahim and Parthasarathy, Nikhil and Evans, Talfan and Beyer, Lucas and Xia, Ye and Mustafa, Basil and others},
  journal={arXiv preprint arXiv:2502.14786},
  year={2025}
}

@misc{torunoğluselamet2026parallelcrosslingualbenchmarkmultimodal,
      title={A Parallel Cross-Lingual Benchmark for Multimodal Idiomaticity Understanding},
      author={Dilara Torunoğlu-Selamet and Dogukan Arslan and Rodrigo Wilkens and Wei He and Doruk Eryiğit and Thomas Pickard and Adriana S. Pagano and Aline Villavicencio and Gülşen Eryiğit and Ágnes Abuczki and Aida Cardoso and Alesia Lazarenka and Dina Almassova and Amalia Mendes and Anna Kanellopoulou and Antoni Brosa-Rodríguez and Baiba Saulite and Beata Wojtowicz and Bolette Pedersen and Carlos Manuel Hidalgo-Ternero and Chaya Liebeskind and Danka Jokić and Diego Alves and Eleni Triantafyllidi and Erik Velldal and Fred Philippy and Giedre Valunaite Oleskeviciene and Ieva Rizgeliene and Inguna Skadina and Irina Lobzhanidze and Isabell Stinessen Haugen and Jauza Akbar Krito and Jelena M. Marković and Johanna Monti and Josue Alejandro Sauca and Kaja Dobrovoljc and Kingsley O. Ugwuanyi and Laura Rituma and Lilja Øvrelid and Maha Tufail Agro and Manzura Abjalova and Maria Chatzigrigoriou and María del Mar Sánchez Ramos and Marija Pendevska and Masoumeh Seyyedrezaei and Mehrnoush Shamsfard and Momina Ahsan and Muhammad Ahsan Riaz Khan and Nathalie Carmen Hau Norman and Nilay Erdem Ayyıldız and Nina Hosseini-Kivanani and Noémi Ligeti-Nagy and Numaan Naeem and Olha Kanishcheva and Olha Yatsyshyna and Daniil Orel and Petra Giommarelli and Petya Osenova and Radovan Garabik and Regina E. Semou and Rozane Rebechi and Salsabila Zahirah Pranida and Samia Touileb and Sanni Nimb and Sarfraz Ahmad and Sarvinoz Nematkhonova and Shahar Golan and Shaoxiong Ji and Sopuruchi Christian Aboh and Srdjan Sucur and Stella Markantonatou and Sussi Olsen and Vahide Tajalli and Veronika Lipp and Voula Giouli and Yelda Yeşildal Eraydın and Zahra Saaberi and Zhuohan Xie},
      year={2026},
      eprint={2601.08645},
      archivePrefix={arXiv},
      primaryClass={cs.CL},
      url={https://arxiv.org/abs/2601.08645},
}

@inproceedings{domhan2022devil,
  title={The devil is in the details: On the pitfalls of vocabulary selection in neural machine translation},
  author={Domhan, Tobias and Hasler, Eva and Tran, Ke M and Trenous, Sony and Byrne, Bill and Hieber, Felix},
  booktitle={Proceedings of the 2022 Conference of the North American Chapter of the Association for Computational Linguistics: Human Language Technologies},
  pages={1861--1874},
  year={2022}
}

@inproceedings{cap2015account,
  title={How to account for idiomatic German support verb constructions in statistical machine translation},
  author={Cap, Fabienne and Nirmal, Manju and Weller, Marion and Im Walde, Sabine Schulte},
  booktitle={Proceedings of the 11th workshop on multiword expressions},
  pages={19--28},
  year={2015}
}

@article{endalie2023deep,
  title={Deep learning-based idiomatic expression recognition for the Amharic language},
  author={Endalie, Demeke and Haile, Getamesay and Taye, Wondmagegn},
  journal={PLoS One},
  volume={18},
  number={12},
  pages={e0295339},
  year={2023},
  publisher={Public Library of Science San Francisco, CA USA}
}

@inproceedings{flor2025survey,
  title={A Survey of Idiom Datasets for Psycholinguistic and Computational Research},
  author={Flor, Michael and Liu, Xinyi and Feldman, Anna},
  booktitle={Proceedings of the 21st Conference on Natural Language Processing (KONVENS 2025): Long and Short Papers},
  pages={90--100},
  year={2025}
}

@article{zeng2021idiomatic,
  title={Idiomatic expression identification using semantic compatibility},
  author={Zeng, Ziheng and Bhat, Suma},
  journal={Transactions of the Association for Computational Linguistics},
  volume={9},
  pages={1546--1562},
  year={2021},
  publisher={MIT Press One Rogers Street, Cambridge, MA 02142-1209, USA journals-info~…}
}

@inproceedings{pickard-etal-2025-semeval,
    title = "{S}em{E}val-2025 Task 1: {A}d{MIR}e - Advancing Multimodal Idiomaticity Representation",
    author = "Pickard, Thomas  and
      Villavicencio, Aline  and
      Mi, Maggie  and
      He, Wei  and
      Phelps, Dylan  and
      Idiart, Marco",
    editor = "Rosenthal, Sara  and
      Ros{\'a}, Aiala  and
      Ghosh, Debanjan  and
      Zampieri, Marcos",
    booktitle = "Proceedings of the 19th International Workshop on Semantic Evaluation (SemEval-2025)",
    month = jul,
    year = "2025",
    address = "Vienna, Austria",
    publisher = "Association for Computational Linguistics",
    url = "https://aclanthology.org/2025.semeval-1.330/",
    pages = "2597--2609",
    ISBN = "979-8-89176-273-2"
}

@inproceedings{jakhotiya2022jarvix,
  title={JARVix at SemEval-2022 Task 2: It Takes One to Know One? Idiomaticity Detection using Zero and One-Shot Learning},
  author={Jakhotiya, Yash and Kumar, Vaibhav and Pathak, Ashwin and Shah, Raj},
  booktitle={Proceedings of the 16th International Workshop on Semantic Evaluation (SemEval-2022)},
  pages={165--168},
  year={2022}
}

@inproceedings{boisson2022cardiffnlp,
  title={Cardiffnlp-metaphor at semeval-2022 task 2: Targeted fine-tuning of transformer-based language models for idiomaticity detection},
  author={Boisson, Joanne and Camacho-Collados, Jose and Anke, Luis Espinosa},
  booktitle={Proceedings of the 16th International Workshop on Semantic Evaluation (SemEval-2022)},
  pages={169--177},
  year={2022}
}

@inproceedings{phelps2024sign,
  title={Sign of the Times: Evaluating the use of Large Language Models for Idiomaticity Detection},
  author={Phelps, Dylan and Pickard, Thomas MR and Mi, Maggie and Gow-Smith, Edward and Villavicencio, Aline},
  booktitle={Proceedings of the Joint Workshop on Multiword Expressions and Universal Dependencies (MWE-UD)@ LREC-COLING 2024},
  pages={178--187},
  year={2024}
}

@inproceedings{wang2025mchirc,
  title={MChIRC: A Multimodal Benchmark for Chinese Idiom Reading Comprehension},
  author={Wang, Tongguan and Wu, Mingmin and Su, Guixin and Su, Dongyu and Hu, Yuxue and Huang, Zhongqiang and Sha, Ying},
  booktitle={Proceedings of the AAAI Conference on Artificial Intelligence},
  volume={39},
  number={24},
  pages={25398--25406},
  year={2025}
}

@article{markchom2025uor,
  title={UoR-NCL at SemEval-2025 Task 1: Using Generative LLMs and CLIP Models for Multilingual Multimodal Idiomaticity Representation},
  author={Markchom, Thanet and Wu, Tong and Huang, Liting and Liang, Huizhi},
  journal={arXiv preprint arXiv:2502.20984},
  year={2025}
}

@article{pickard2025semeval,
  title={SemEval-2025 Task 1: AdMIRe--Advancing Multimodal Idiomaticity Representation},
  author={Pickard, Thomas and Villavicencio, Aline and Mi, Maggie and He, Wei and Phelps, Dylan and Idiart, Marco},
  journal={arXiv preprint arXiv:2503.15358},
  year={2025}
}

@article{khan2025evaluating,
  title={Evaluating Large Language Models on Urdu Idiom Translation},
  author={Khan, Muhammad Farmal and Akter, Mousumi},
  journal={arXiv preprint arXiv:2510.17460},
  year={2025}
}

@article{phelps2022sample,
  title={Sample Efficient Approaches for Idiomaticity Detection},
  author={Phelps, Dylan and Fan, Xuan-Rui and Gow-Smith, Edward and Madabushi, Harish Tayyar and Scarton, Carolina and Villavicencio, Aline},
  journal={MWE 2022},
  pages={105},
  year={2022}
}

@article{xing2024survey,
  title={A survey of efficient fine-tuning methods for vision-language models—prompt and adapter},
  author={Xing, Jialu and Liu, Jianping and Wang, Jian and Sun, Lulu and Chen, Xi and Gu, Xunxun and Wang, Yingfei},
  journal={Computers \& Graphics},
  volume={119},
  pages={103885},
  year={2024},
  publisher={Elsevier}
}

@inproceedings{radford2021learning,
  title={Learning transferable visual models from natural language supervision},
  author={Radford, Alec and Kim, Jong Wook and Hallacy, Chris and Ramesh, Aditya and Goh, Gabriel and Agarwal, Sandhini and Sastry, Girish and Askell, Amanda and Mishkin, Pamela and Clark, Jack and others},
  booktitle={International conference on machine learning},
  pages={8748--8763},
  year={2021},
  organization={PmLR}
}

@article{gao2024clip,
  title={Clip-adapter: Better vision-language models with feature adapters},
  author={Gao, Peng and Geng, Shijie and Zhang, Renrui and Ma, Teli and Fang, Rongyao and Zhang, Yongfeng and Li, Hongsheng and Qiao, Yu},
  journal={International Journal of Computer Vision},
  volume={132},
  number={2},
  pages={581--595},
  year={2024},
  publisher={Springer}
}

@article{jin2025efficient,
  title={Efficient multimodal large language models: A survey},
  author={Jin, Yizhang and Li, Jian and Gu, Tianjun and Liu, Yexin and Zhao, Bo and Lai, Jinxiang and Gan, Zhenye and Wang, Yabiao and Wang, Chengjie and Tan, Xin and others},
  journal={Visual Intelligence},
  volume={3},
  number={1},
  pages={27},
  year={2025},
  publisher={Springer}
}

\appendix

\section{Appendix}

Table~\ref{tab:per_language_subtasks} summarises the official Codabench blind test results for our submission \texttt{nikoniko}. For Subtask~A (image+text), our system achieved an average Top\textsubscript{1} accuracy of 0.35 and an NDCG of 0.73 across 15 languages. For Subtask~B (text only), the corresponding scores were 0.32 Top\textsubscript{1} and 0.71 NDCG. In both subtasks, our run ranked fifth on the Codabench leaderboard at evaluation time. The per language breakdown shows clear variation across the multilingual blind test. For Subtask~A, the best performing languages are Portuguese Brazil (PT-BR) (0.46), PT (0.42), and Norwegian (0.42), while Spanish Ecuador (ES-EC) (0.17), Uzbek (UZ) (0.29), and Georgian (KA) (0.27) emerge as the most challenging. Despite these differences in Top\textsubscript{1} accuracy, NDCG scores remain consistently high in the 0.69 to 0.76 range, with an overall average of 0.73.

\begin{table}[hpt!]
\centering
\scriptsize
\setlength{\tabcolsep}{2pt}
\resizebox{\columnwidth}{!}{%
\begin{tabular}{lcccc}
\toprule
\multirow{2}{*}{Language} 
  & \multicolumn{2}{c}{Subtask A (img+txt)} 
  & \multicolumn{2}{c}{Subtask B (text)} \\
\cmidrule(lr){2-3} \cmidrule(lr){4-5}
  & Top$_1$ (\%) & NDCG@5 & Top$_1$ (\%) & NDCG@5 \\
\midrule
Chinese (ZH)           & 35.0 & 0.72 & 28.0 & 0.69 \\
Georgian (KA)          & 27.0 & 0.69 & 28.0 & 0.68 \\
Greek (EL)             & 36.0 & 0.72 & 28.0 & 0.71 \\
Igbo (IG)              & 33.0 & 0.73 & 29.0 & 0.69 \\
Kazakh (KK)            & 33.0 & 0.74 & 28.0 & 0.72 \\
Norwegian (NO)         & 42.0 & 0.75 & 41.0 & 0.75 \\
Portuguese BR (PT-BR)  & 46.0 & 0.76 & 37.0 & 0.75 \\
Portuguese (PT)  & 42.0 & 0.75 & 37.0 & 0.73 \\
Russian (RU)           & 40.0 & 0.73 & 39.0 & 0.74 \\
Serbian (SR)           & 39.0 & 0.74 & 31.0 & 0.71 \\
Slovak (SK)            & 38.0 & 0.73 & 35.0 & 0.73 \\
Slovenian (SL)         & 40.0 & 0.75 & 37.0 & 0.74 \\
Spanish EC (ES-EC)     & 17.0 & 0.66 & 19.0 & 0.67 \\
Turkish (TR)           & 34.0 & 0.71 & 32.0 & 0.70 \\
Uzbek (UZ)             & 29.0 & 0.71 & 32.0 & 0.71 \\
\midrule
Macro average          & 35.4 & 0.72 & 32.1 & 0.72 \\
\bottomrule
\end{tabular}%
}
\caption{Per language performance on the multilingual blind test. Subtask~A uses images and captions; Subtask~B is text only.}
\label{tab:per_language_subtasks}
\end{table}


\end{document}